\documentclass[review]{elsarticle}

\usepackage{amsmath}
\usepackage{lineno,hyperref}
\usepackage{graphicx}
\modulolinenumbers[5]

\graphicspath{}

\begin{document}
\begin{frontmatter}

\title{An (almost) instant brain atlas segmentation for large-scale studies}

\author[mrn,ece]{Alex Fedorov\corref{cor1}}
\ead{afedorov@mrn.org}
\author[mrn]{Eswar Damaraju}

\author[mrn,ece]{Vince Calhoun}
\author[mrn]{Sergey Plis\corref{cor1}}
\ead{splis@mrn.org}

\cortext[cor1]{Corresponding authors at: 1101 Yale Blvd NE, Albuquerque, NM 87106, USA}

\address[mrn]{The Mind Research Network, Albuquerque, United States}
\address[ece]{Department of Electrical and Computer Engineering, University of New Mexico, Albuquerque, United States}

\begin{abstract}
  Large scale studies of group differences in healthy controls and patients and screenings for early stage disease prevention programs require processing and analysis of extensive multisubject datasets.
  Complexity of the task increases even further when segmenting structural MRI of the brain into an atlas with more than 50 regions.
  Current automatic approaches are time-consuming and hardly scalable; they often involve many error prone intermediate steps and don't utilize other available modalities.
  To alleviate these problems, we propose a feedforward fully convolutional neural network trained on the output produced by the state of the art models.
  Incredible speed due to available powerful GPUs neural network makes this analysis much easier and faster (from $>10$ hours to a minute).
  The proposed model is more than two orders of magnitudes faster than the state of the art and yet as accurate.
  We have evaluated the network's performance by comparing it with the state of the art in the task of differentiating region volumes of healthy controls and patients with schizophrenia on a dataset with 311 subjects.
  This comparison provides a strong evidence that speed did not harm the accuracy.
  The overall quality may also be increased by utilizing multi-modal datasets (not an easy task for other models) by simple adding more modalities as an input.
  Our model will be useful in large-scale studies as well as in clinical care solutions, where it can significantly reduce delay between the patient screening and the result.
\end{abstract}

\end{frontmatter}

\section{Introduction}

Structural magnetic resonance imaging (sMRI) is an essential tool for the clinical care of patients providing details about the anatomical structure of the brain. Segmenting a structural MRI is an important processing step to provide regional tissue volumes and enables subsequent inferences about tissue changes in development, aging and as well as disease. With increasing number of large scale studies and screening programs to detect signs of the disease in early stages it will affect the number of medical images have to be analyzed. Therefore we need fast and accurate enough tools for big data oriented medical research. It will also provide significant benefits for radiologists, neurologists and patients in the medical settings.

Manual labeling is currently gold standard for all kinds of segmentation of medical images. But labeling time cost of an one entire high resolution structural MRI of brain image is about week for an expert which is not acceptable in large scale studies. Furthermore, labeling is usually performed on slices which makes it more consistent along the slice direction. While multi-modal images (T1-weighted, T2-weighted, FLAIR) can provide better tissue contrast to facilitate segmenting, manually performing the same is not only tedious but also more complicated. Therefore, automated methods for segmenting and parcellating the sMRI images are very popular today. Most common tools with strong support community and high success rates~\cite{MIKHAEL2017} in internal validation are FreeSurfer~\cite{Desikan2006968, Fischl01012004, destrieux2010automatic}, BrainSuite~\cite{shattuck2002brainsuite}, BrainVISA~\cite{Geffroy11}. These softwares primarily employ probabilitic methods with priors to solve the problem.

Recently, with the success of end-to-end deep learning approaches using convolutional neural networks (CNNs) identification and segmentation of visual imagery, several new common approaches based on fully convolutional networks~\cite{long2015fully}, U-Net~\cite{ronneberger2015u, cciccek20163d}, Pyramid-long short term memory (LSTM)~\cite{stollenga2015parallel}, dilated convolutions~\cite{yu2015multi} have been developed for segmentation of medical images. These methods can combine multiple sMRI image modalities as input to use different tissue contrast and have the ability to produce fast and accurate prediction. However these methods require substantial amounts of training data, which is often difficult to obtain due to the lack of enough labeled data in medical field.

Fully convolutional networks are a type of CNN where fully connected layers have been replaced by convolutional layers. It allows to predict specific area or volume instead of running CNN with fully connected layers pixel-wise or voxel-wise. This approach has recently been used for segmentation of subcortical regions~\cite{DOLZ2017}.

U-Net is a generative ladder network consisting of encoder and decoder. This kind of network has been used for visual cortex parcellation~\cite{7950666}. In this article, the authors worked with 2D slices and also added additional encoder network for atlas prior, which increases the number of parameters and requires some prior information.

Pyramid-LSTM is a deep neural network based on convolutional LSTM (C-LSTM) layers. To work with 3D volumes it requires 6 C-LSTM layers working with 2D images to scan volume in 6 directions after which it combines output of these C-LSTMs and uses pixel-wise fully connected layers to provide final prediction. This architecture has been used for brain tissue segmentation in MRI images and segmentation of neuronal processes in electron microscopy images.

From simple Fully-convolutional to Pyramid-LSTM, the complexity of neural network architecture is increased with skip-connections in U-Net or C-LSTMs in Pyramid-LSTM. But our MeshNet~\cite{fedorov2017end} has a simple feedforward fully convolutional arhitecture with dilated kernels~\cite{yu2015multi} and we successfully demonstrated its use in gray and white matter segmentation. But this task is simple compared to whole brain atlas segmentation containing more than 50 labels. The atlas segmentation enables more detailed comparing of regional growth curves in gray matter densities \cite{sowell2003mapping,tamnes2009brain} or differences in regional gray matter volumes between healthy controls and patients, for example autism \cite{redcay2005brain}, schizophrenia \cite{van2016subcortical} and \cite{okada2016abnormal}. 

In this work we extend our MeshNet model to perform atlas segmentation task with 50 cortical, subcortical and cerebellar labels obtained from Freesurfer on open Human Connectome Project (HCP) data and subsequently evaluate the trained models performance in differentiating regional volumes between healthy controls and patients with schizophrenia compared to those obtained using FreeSurfer using a previously published multi-site functional bioinformatics informatics research (fBRIN) data.

\section{Materials and Methods}

\subsection{Feedforward and Convolutional Neural Networks}

Feedforward Neural Network or multilayer perceptron (MLPs) are classic model to approximate some function, for example, classifier, $y=F(\mathbf{x})$ maps input $\mathbf{x}$ to output $y$. Mapping $F$ can be represented as $n$-layer network $F(x) = f^n(f^{n-1}(...(f^1(\mathbf{x}))))$. Each function $f^i$ is can be written by equation~\ref{eq:layer} as linear model $\mathbf{W}_i^\top\mathbf{x}+b_i$ with weights $\mathbf{W}_i$ and bias $b_i$ and some nonlinear function $g$, known as activation function.

\begin{equation} 
f^i(\mathbf{x}) = g(\mathbf{W}_i^\top f^{i-1}(\mathbf{x})+b_i)
\label{eq:layer}
\end{equation}
where $f^0(\mathbf{x}) = \mathbf{x}$.

In case of Convolutional Neural Networks instead of matrix multiplication $\mathbf{W}_i^\top\mathbf{x}$ we are using convolution~\ref{eq:conv} with some kernel $\mathbf{W}_i$.

\begin{equation}
f^i(\mathbf{x}) = g(\mathbf{W}_i\ast f^{i-1}(\mathbf{x}) + b_i)
\label{eq:conv}
\end{equation}

Function $f^i$ can be also defined by pooling as max pooling~\cite{ZhouChellappa}, normalization as batch normalization~\cite{ioffe2015batch} or dropout~\cite{srivastava2014dropout} layers.

Convolutional neural networks has some advantages~\cite{Goodfellow-et-al-2016} compared to MLP. First advantage is local or sparse connectivity. If in MLPs we are connecting every input with every output, here we are applying kernel to only small region of input defined by kernel size, but in deeper layers neurons still indirectly connected to most part of input. How large neuron is connected to input is determined by receptive field, depends on neural network hyperparameters and architecture. Overall local connectivity reduces number of parameters, computational complexity and memory requirements. Second advantage comes from that we are using same kernel over all input, known as parameter sharing. Third advantage is equivariance to translation due to convolutional nature. Other transformations as rotation and scaling requires additional techniques.

\subsection{Convolutional Neural Networks for dense prediction}

\subsubsection{Fully convolutional networks}

\begin{figure}[ht]
\centering
\includegraphics[scale=.4]{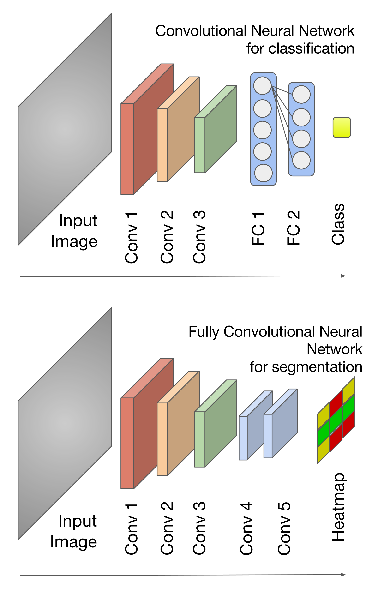}
\caption{Convolutional and Fully Convolutional Neural Networks}
\label{fig:conv_and_fconv}
\end{figure}

Fully convolutional neural networks (FCNN)~\cite{long2015fully} are the type of convolutional neural networks where fully connected (FC) layers is replaced by convolutional layers. Fig. \ref{fig:conv_and_fconv} visually explains difference between them. Output of FCNN is heatmap, where in 2D case each pixel suggests specific label, whenever neural network with FC layers produces just one label. This kind of networks is suitable for end-to-end learning and pixel-to-pixel segmentation.

\begin{figure}[ht]
\centering
\includegraphics[scale=.6]{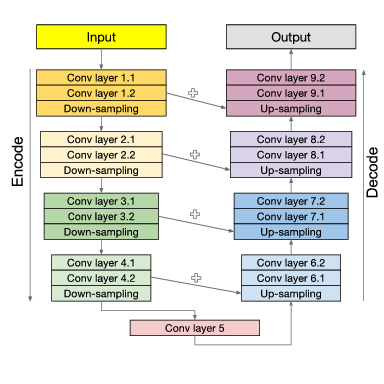}
\caption{U-Net}
\label{fig:u-net}
\end{figure}

\subsubsection{U-Net}
U-Net is advancement of fully convolutional networks by adding decoder part to encoder part and connections between them. Example of U-Net architecture is shown on Fig.~\ref{fig:u-net}. Encoder uses convolutional layers to shrink image information to compact representation and then decoder using upsampling layers and concatenation connections restores output to initial size of input. Connections are needed to connect representation of decoder with encoder from higher resolution features to produce upsampling, or by another words it allows to flow higher resolution information. U-Net, compared to FCNN, allows to produce segmentation with same size as input image.

\subsubsection{Convolutional Neural Network with dilated kernels}

Discrete volumetric convolution can be written as
\begin{equation}
(k*f)_{(x,y,z)} = \sum_{\bar{x}=-a}^a \sum_{\bar{y}=-b}^b \sum_{\bar{z}=-c}^c k(\bar{x},\bar{y},\bar{z})f(x-\bar{x},y-\bar{y},z-\bar{z}),
\label{eq:3dconv}
\end{equation}
where $a$, $b$, $c$ are kernel bounds on $x$, $y$ and $z$ axis respectively and $(x,y,z)$ is the point at which we compute the convolution.
We can extend it to make L-Dilated convolution as:
\begin{equation}
(k*_lf)_{(x,y,z)}\!=\!\sum_{\bar{x}=-a}^a \sum_{\bar{y}=-b}^b \sum_{\bar{z}=-c}^c k(\bar{x},\bar{y},\bar{z})f(x-l\bar{x},y-l\bar{y},z-l\bar{z}),
\label{eq:dil3dconv}
\end{equation}
where $l$ is dilation factor, $*_l$ is $l$-dilated convolution.

First dilated convolutions using CNNs was introduced in article~\cite{yu2015multi}. Using L-Dilated convolution in CNNs can allow drastically increase receptive field of the neuron, but preserve small amount of parameters. In traditional CNNs to have same receptive field we possibly have to increase depth of neural network or change the kernel size. But this will generate an increased number of parameters to fit, especially in 3D case. Dilated kernels also allows to control receptive field of a neuron while preserving the same number of parameters and depth of neural network.

\subsubsection{MeshNet}

MeshNet~\cite{fedorov2017end} is feed-forward convolutional neural network with dilated kernels. Every layer of this network create same size feature maps and output in the end as input. Architecture has been inspired by context module of~\cite{yu2015multi} and has been extended to 3D case. 

\begin{figure}[ht]
\centering
\includegraphics[scale=.6]{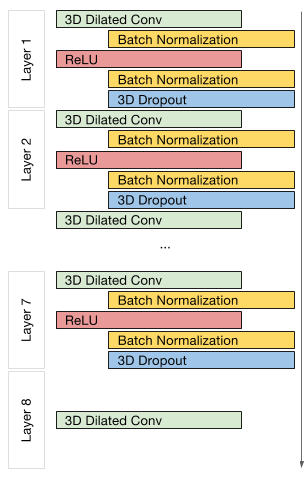}
\caption{MeshNet architecture. Shifted blocks with Batch Normalization and 3D dropout means possible location for layer. But Batch Normalization can be just before or after ReLU in every layer besides last 8th layer.}
\label{fig:meshnet}
\end{figure}

In our previous model~\cite{fedorov2017end} we replaced 1D dropout by volumetric 3D dropout, because volumetric feature maps activations are strongly correlated due nature of images~\cite{tompson2015efficient}. We also tried to use batch normalization before and after activation function for faster convergence. As activation function we are using ReLU. For weight initialization we have tried identity initialization from~\cite{yu2015multi} and also xavier initialization~\cite{glorot2010understanding}. To get prediction we are using LogSoftMax activation function after last layer and afterwards maximum value to determine label of the voxel. But overall architecture is the same and scheme of architecture is shown on a Fig.~\ref{fig:meshnet}. To train neural network we are using variation of stochastic gradient descent Adam~\cite{kingma2014adam} to automatically adjust learning rate. For a loss function we have chosen categorical cross-entropy function. 

\begin{figure}[ht]
\centering
\includegraphics[width=\linewidth]{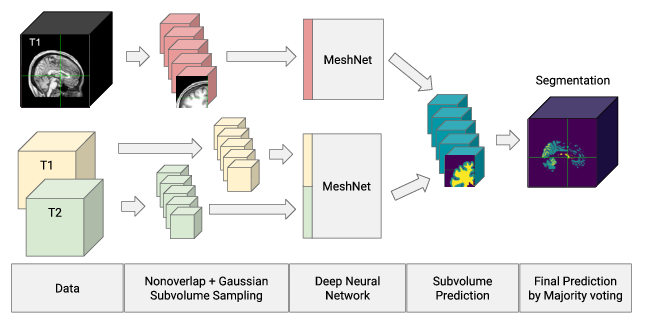}
\caption{MeshNet workflow}
\label{fig:workflow}
\end{figure}

Our workflow with Meshnet is shown in Figure~\ref{fig:workflow}. Training worklfow consist from nonoverlap and gaussian subvolume sampling of input T1 or T2 volumes, feeding them to MeshNet, obtaining subvolume prediction by taking argmax of LogSoftMax class channel in every voxel. Final segmentation for testing is obtained by voxel-wise majority voting on accumulated prediction from subvolumes. But last steps as argmax on LogSoftMax and majority voting can be modified by more complicated approaches.

For atlas segmentation we reduced volumetric shape of input to $38^3$ from $68^3$ as we used in our previous work~\cite{fedorov2017end} to fit model in GPU memory and have more features maps in every layer due to increased complexity of the task. Final hyperparameters of the model is shown in Table~\ref{tab:meshnettable_atlas}. 

\begin{table}[]
\centering
\caption{Detailed hyperparameters of MeshNet for atlas segmentation with subvolume side length 38. $k^3$ means $k\times k\times k$. $M$ is number of input modalities, $M$ equal to 1 or 2 if we are using T1 or T1 and T2 respectively. The output of th 8th layer is 50 feature maps, equal to number of regions in our atlas. 3D DC stands for volumetric dilated convolution, BN --- Batch Normalization, ReLU --- Rectified Linear Unit. *In the 7th layer models can have 3D dropout with $p=0.125$ and $p=0.25$ for additional regularization or don't have it. }
\label{tab:meshnettable_atlas}
\scalebox{0.7}{\begin{tabular}{lllllll}
\hline
  & Layer & Kernel & Input & Output & Padding & Dilation \\ \hline
1 & 3D DC + BN + ReLU & $3^3$ & $M \times 38^3$ & $71 \times 38^3$& 1 & 1 \\ 
2 & 3D DC + BN + ReLU & $3^3$ & $71 \times 38^3$& $71 \times 38^3$& 1 & 1 \\ 
3 & 3D DC + BN + ReLU & $3^3$ & $71 \times 38^3$& $71 \times 38^3$& 1 & 1 \\ 
4 & 3D DC + BN + ReLU & $3^3$ & $71 \times 38^3$& $71 \times 38^3$& 2 & 2 \\ 
5 & 3D DC + BN + ReLU & $3^3$ & $71 \times 38^3$& $71 \times 38^3$& 4 & 4 \\ 
6 & 3D DC + BN + ReLU & $3^3$ & $71 \times 38^3$& $71 \times 38^3$& 8 & 8 \\ 
7 & 3D DC + BN + ReLU + 3D Dropout(p)* & $3^3$ & $71 \times 38^3$& $71 \times 38^3$& 1 & 1 \\ 
8 & 3D DC & $3^3$ & $71 \times 38^3$ & $50 \times 38^3$ & 0 & 1 \\ \hline
\multicolumn{5}{c}{Receptive Field} & \multicolumn{2}{l}{$37^3$} \\ \hline
\multicolumn{5}{c}{Overall number of parameters} & \multicolumn{2}{l}{825567} \\ \hline
\end{tabular}}
\end{table}

\subsection{Metrics}
For measuring the performance of the models we are calculating DICE coefficient~\cite{taha2015metrics} for measuring spatial overlap and Average Volume Difference (AVD)~\cite{mendrik2015mrbrains} for validating volume values. DICE coefficient is defined as 
\begin{equation} \label{eq:dice}
  DICE = \frac{2 |P \cap G|}{|P| + |G|} = \frac{2TP}{2TP + FN + FP},
\end{equation}
where $P$ is the prediction, $G$ --- the ground truth, $TP$ --- the true positive value, $FP$ --- the false positive value, $FN$ --- the false negative value. The Average Volume Difference is defined as
\begin{equation} \label{eq:avd}
  AVD = \frac{|V_p \cap V_g|}{V_g},
\end{equation}
where $V_p$ is the prediction volume by model and $V_g$ is the ground truth volume. This metric is sensitive to point positions.

We are also using Macro-averaging~\cite{zhang2014review} approach for comparing overall performance of prediction for different number of subvolumes.
Macro-averaging for metric $B$ (DICE, AVD) is defined as:
\begin{equation} \label{eq:macro-averaging}
  B_{macro} = \frac{1}{C} \sum_{c=1}^C B_c,
\end{equation}
where $C$ is number of classes and $B_c$ --- metric $B$ for class $c$.

\subsection{Datasets}

\subsubsection{Human Connectome Project dataset}

The Human Connectome Project (HCP)~\cite{VANESSEN201362} is an open-access dataset with multi-modal brain imaging data for healthy young-adult subjects. From this dataset we are using T1 and T2 of $887$ subjects with slice thickness $0.7 \times 0.7 \times 0.7$. These T1 and T2 have been resliced to $1.0 \times 1.0 \times 1.0$ and padded with zeros to a volume with dimensions $256 \times 256 \times 256$, also min-max normalization has been applied. Dataset has been splitted on $770$ volumes for training, $27$ --- validation and $100$ --- testing.

\subsubsection{Function Biomedical Informatics Research Network dataset}

The Function Biomedical Informatics Research Network (FBIRN) dataset~\cite{KEATOR20161074} is an open-access dataset includes multi-modal structural MRI, fMRI, DTI, behavioral data and clinical, demographic assessments. We used T1-weighted MRI scans from FBIRN Phase III and atlas segmentation labels obtained using FreeSurfer 5.3 as in HCP. T2-weighted MRI scans had just $32$ axial slices therefore we decided not to use them. Data have been collected on 3.0 Tesla MRI scanners on GE and Siemens platforms at $7$ different sites. Overall dataset consist from $348$ subjects of which $171$ are patients with schizophrenia and $177$ --- healthy subjects. Subjects are in the age between $16$-$62$. $238$ of the subject are males, $86$ --- females and $24$ --- unknown. All data has been resliced to $1.0 \times 1.0 \times 1.0$ thickness,  padded with zeros to $256 \times 256 \times 256$ and min-max normalized to be the same as HCP dataset. For fine-tuning we used 7 subjects (1 subject per site) as training dataset and another 7 subjects as validation dataset therefore $334$ subjects are testing set.

\subsection{Statistical meta-analysis} \label{ssec:stat-meta}

In structural studies voxel-based morphometry analysis of MRI images have been used to compare regions of interests (ROIs) in the brain between patient and healthy controls, to find ROI with possible abnormalities. Some previous works are \cite{van2016subcortical} and \cite{okada2016abnormal}, related to schizophrenia as FBIRN dataset. Therefore to compare FreeSurfer with proposed method we perform similar meta-analysis.

First, we defined univariate linear regression model for prediction of ROI's volume $V_{roi}$:
\begin{multline}
V_{roi} = age + age^2 + gender + V_{brain} + gender:age + gender:V_{brain} + \\
  site_1 + site_2 + site_3 + site_4 + site_5 + site_6,
\end{multline}
where $age$ is value between $18$ to $62$, $gender$ is $0$ for male and $1$ --- female, $V_{brain}$ is sum of all brain ROIs, $gender:age$ and $gender:V_{brain}$ are interactions of $age$ and $V_{brain}$ with $gender$ and $site_i, i = 1,..,6$ is one-hot encoding of $7$ sites. Volume is measured in number of voxels. This step is adjusting ROI's volume for age, gender, brain volume and sites.

Secondly, we calculated residuals between true value of ROI's volume and predicted. Then linear model for repeated measurements ANOVA is defined:
\begin{equation}
V^{roi}_{residual} = method + label + method:label,
\end{equation}
where $V^{roi}_{residual}$ is residual for ROI after first step, $method$ is method used to compute $V_{roi}$: FreeSurfer or MeshNet, $label$ is healthy subject or patient with Schizophrenia, $method:label$ --- interaction of $method$ and $label$.

To compare methods we inspected significance of covariates $label$ and $method:label$ and also Cohen's d was computed on ROI's volume residuals after first step between groups.

\section{Results}

First we summarize performance of MeshNet on HCP dataset, then present results of a fine-tuned for FBIRN dataset model, trained before on HCP dataset, and statistical comparison of MeshNet with FreeSurfer.

\subsection{Segmentation quality and speed performance}

\subsubsection{Speed performance}

\begin{figure}[h]
\centering
\includegraphics[width=\textwidth, keepaspectratio]{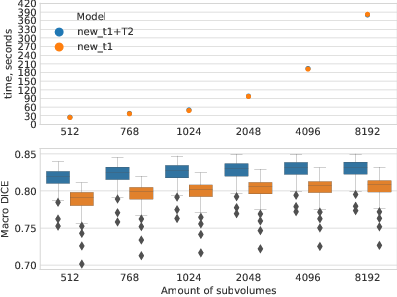}
\caption{Prediction time in seconds and overall segmentation quality measured with macro DICE as a function of the amount of subvolumes.  Scores were calculated on 100 testing subjects. For macro DICE higher is better. Time is measured on GeForce Titan X Pascal 12Gb.}
\label{fig:speed}
\end{figure}

Since our model predicts for $38 \times 38 \times 38$ subvolumes, we measured how long it takes to create whole $256 \times 256 \times 256$ MRI image segmentation and how well it performs with the amount of subvolumes of $512$, $768$, $1024$, $2048$, $4096$ and $8192$. Considering Figure~\ref{fig:speed}, model can offer acceptable segmentation with $1024$ subvolumes running in average 48 seconds for T1 model and in average 50 seconds for T1 and T2 model  on GeForce Titan X Pascal 12Gb.

\subsubsection{Segmentation quality on HCP}  \label{ssec:qual_hcp}

\begin{figure}[p]
\centering
\includegraphics[width=\textwidth, keepaspectratio]{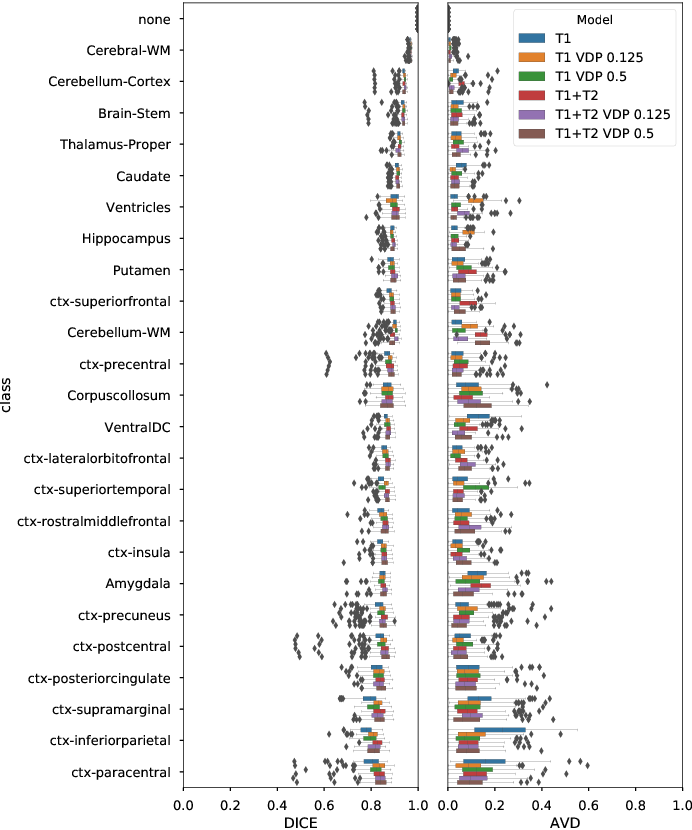}
\caption{DICE and AVD scores evaluated on $100$ testing subjects from HCP dataset for first $25$ from $50$ regions, sorted by maximum region-wise DICE score of T1+T2 model. DICE and AVD were measured on segmentation of whole MRI image $256\times  256\times  256$ with sampling of $1024$ $38\times  38\times 38$ subvolumes. For DICE to the right (higher values) is better and for AVD to the left (lower values) is better.}
\label{fig:metrics_part1}
\end{figure}

\begin{figure}[p]
\centering
\includegraphics[width=\textwidth, keepaspectratio]{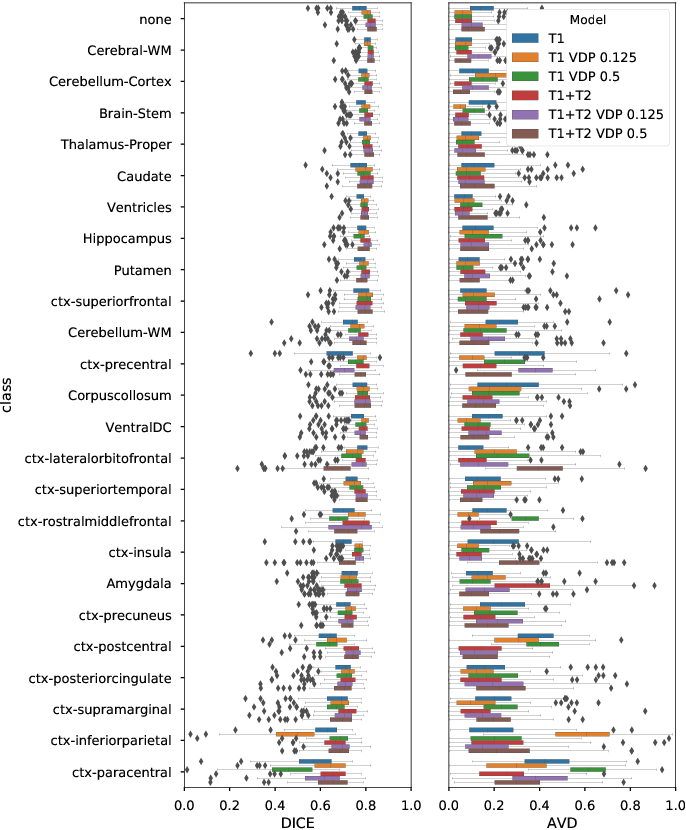}
\caption{DICE and AVD scores evaluated on $100$ testing subjects from HCP dataset for last $25$ from $50$ regions, sorted by maximum region-wise DICE score of T1+T2 model. DICE and AVD were measured on segmentation of whole MRI image $256\times  256\times  256$ with sampling of $1024$ $38\times  38\times 38$ subvolumes. For DICE to the right (higher values) is better and for AVD to the left (lower values) is better.}
\label{fig:metrics_part2}
\end{figure}

\begin{figure}[p]
\centering
\includegraphics[width=\textwidth, keepaspectratio]{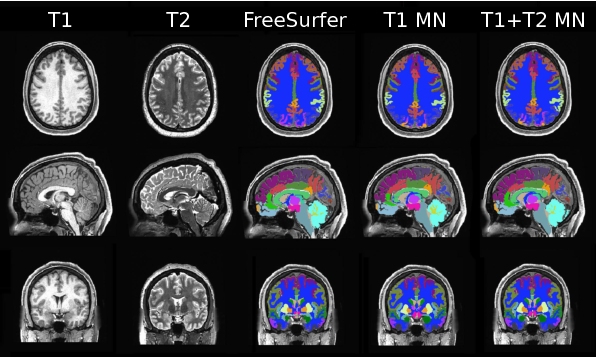}
\caption{First segmentation example.}
\label{fig:hcp_brain1}
\end{figure}

\begin{figure}[p]
\centering
\includegraphics[width=\textwidth, keepaspectratio]{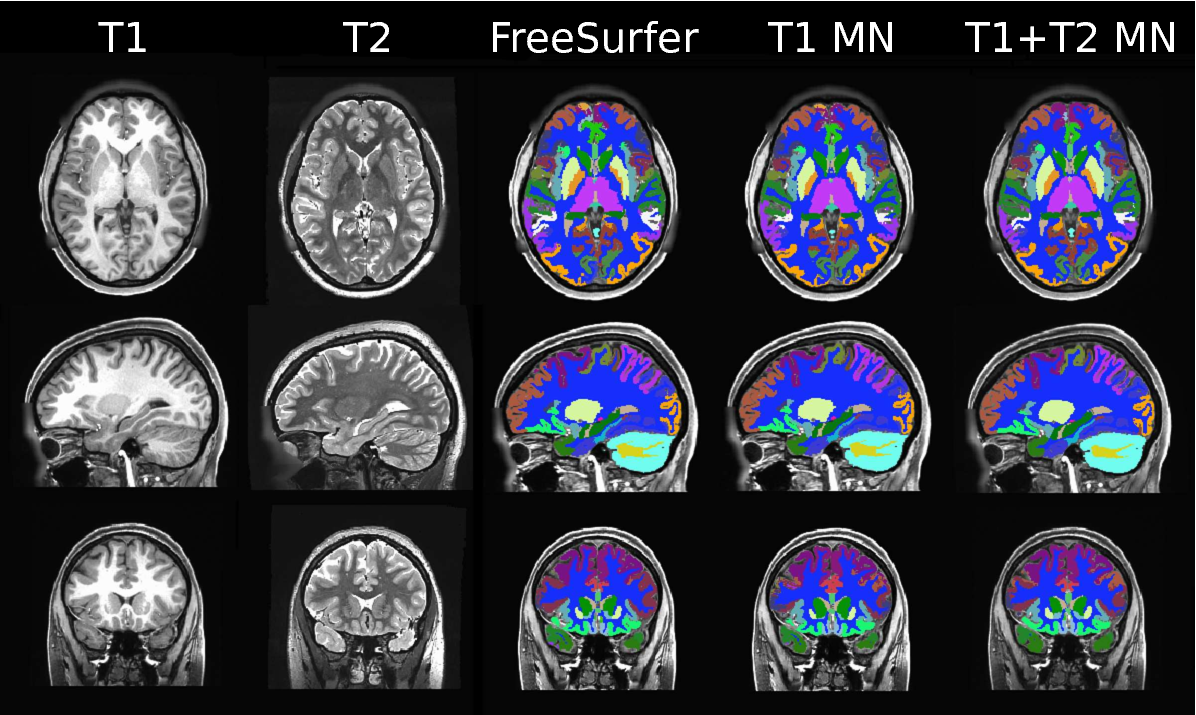}
\caption{Second segmentation example.}
\label{fig:hcp_brain2}
\end{figure}

Segmentation quality on $100$ testing subject from HCP dataset are shown in Figure~\ref{fig:metrics_part1} and Figure~\ref{fig:metrics_part2}. DICE and AVD scores were measured on whole MRI segmentation, obtained by sampling of $1024$ subvolumes with shape $38\times  38\times 38$. The model with T1 and T2 as input shows best overall performance. Additional regularization of T1 and T2 model with volumetric dropout before last layer in some classes shows better scores than nondropout model, but model can, for example, in precentral cortex with probability of dropout 0.125 or lateral orbitofrontal cortex with probability of dropout 0.25 can perform poorer than nondropout T1 model, therefore nondropout more stable in overall performance. Same is applicable for T1 model with dropout. Segmentation examples prepared using AFNI~\cite{cox1996afni} for T1 and T1, T2 nondropout models are shown in Figures~\ref{fig:hcp_brain1} and \ref{fig:hcp_brain2}.

\subsubsection{Segmentation quality on FBIRN} \label{ssec:qual_fbirn}

\begin{figure}[p]
\centering
\includegraphics[width=\textwidth, keepaspectratio]{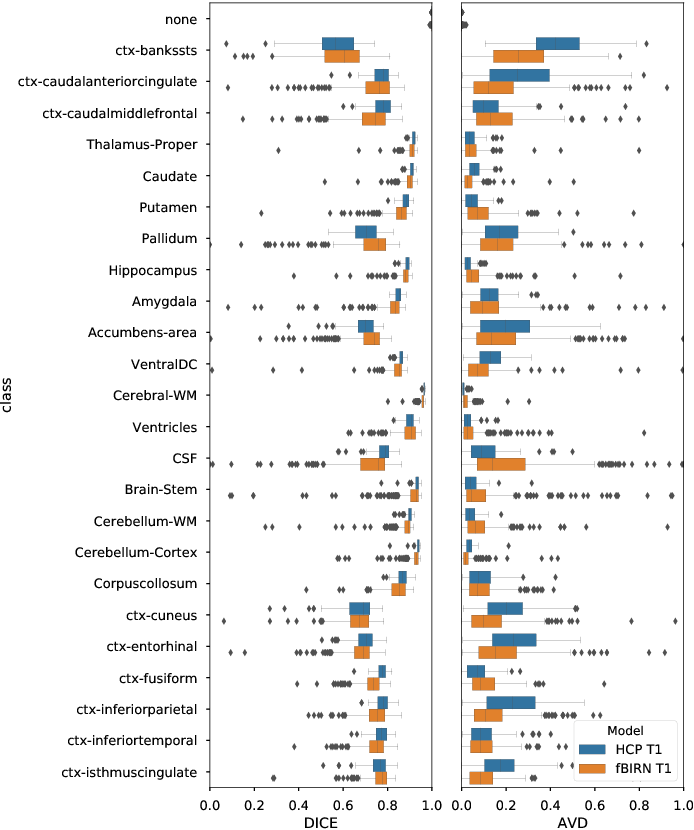}
\caption{DICE and AVD scores evaluated on $100$ testing subject from HCP and on $334$ testing subject from FBIRN dataset for 25 regions from 50. DICE and AVD were measured on segmentation of whole MRI image $256\times  256\times  256$ with sampling of $1024$ $38\times  38\times 38$ subvolumes. For DICE to the right (higher values) is better and for AVD to the left (lower values) is better.}
\label{fig:fbirn_metrics_part1}
\end{figure}

\begin{figure}[p]
\centering
\includegraphics[width=\textwidth, keepaspectratio]{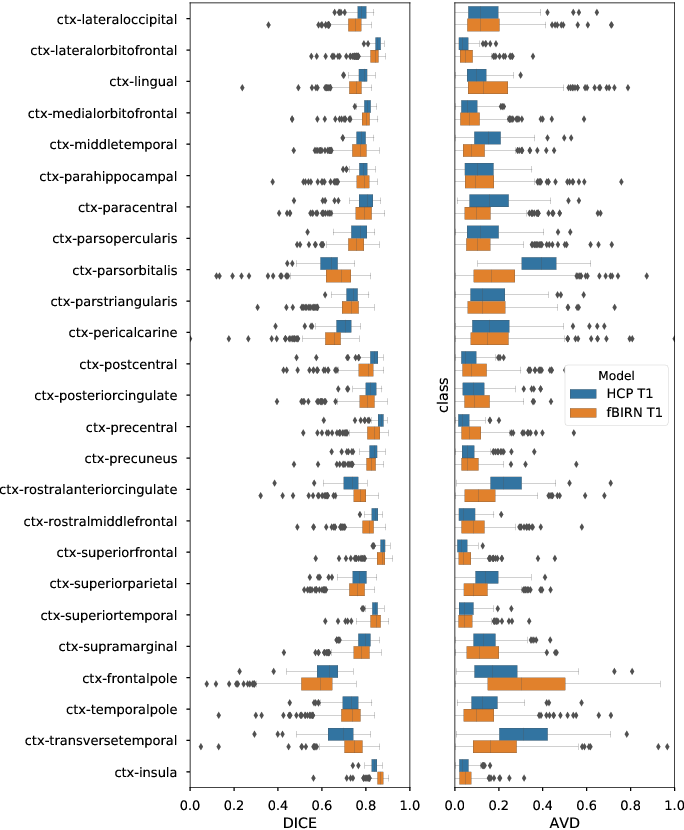}
\caption{DICE and AVD scores evaluated on $100$ testing subject from HCP and on $334$ testing subject from FBIRN dataset for another 25 regions from 50. DICE and AVD were measured on segmentation of whole MRI image $256\times  256\times  256$ with sampling of $1024$ $38\times  38\times 38$ subvolumes. For DICE to the right (higher values) is better and for AVD to the left (lower values) is better.}
\label{fig:fbirn_metrics_part2}
\end{figure}

\begin{figure}[p]
\centering
\includegraphics[scale=0.45, keepaspectratio]{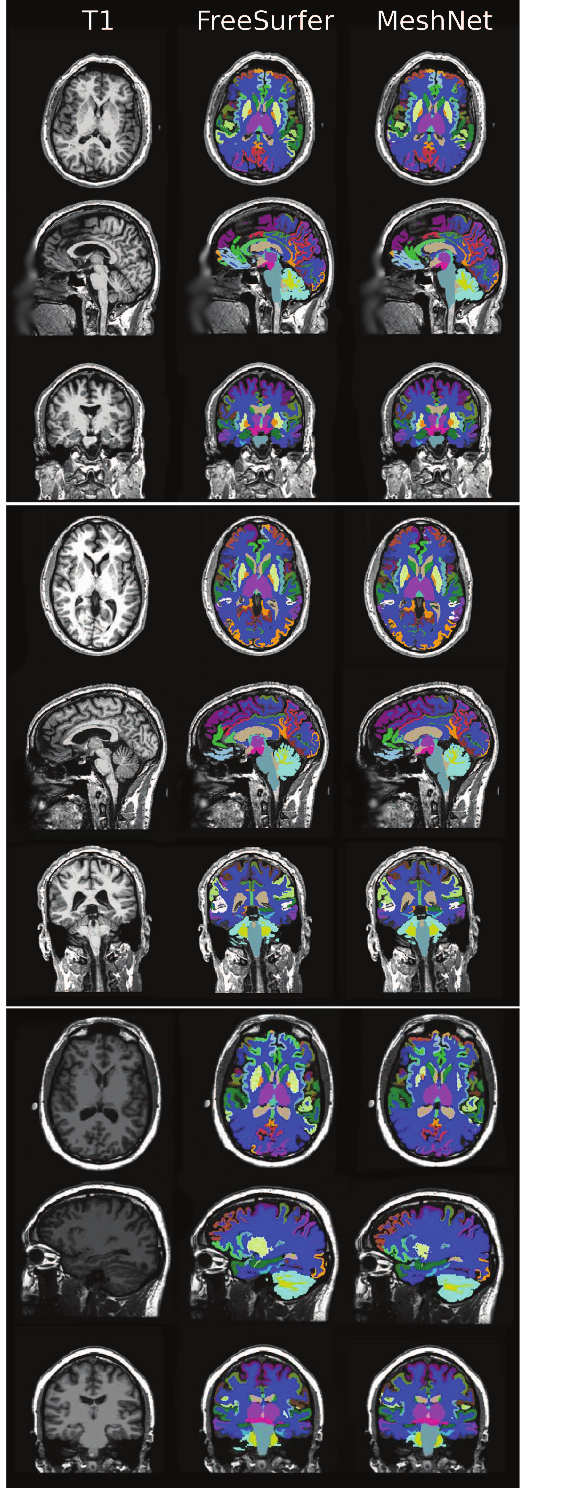}
\caption{Segmentation example for 1 healthy subject and 2 patients with schizophrenia from FBIRN dataset. Face was blurred manually after performing segmentation.}
\label{fig:fbirn_brains}
\end{figure}

\begin{figure}[p]
\centering
\includegraphics[width=\textwidth, keepaspectratio]{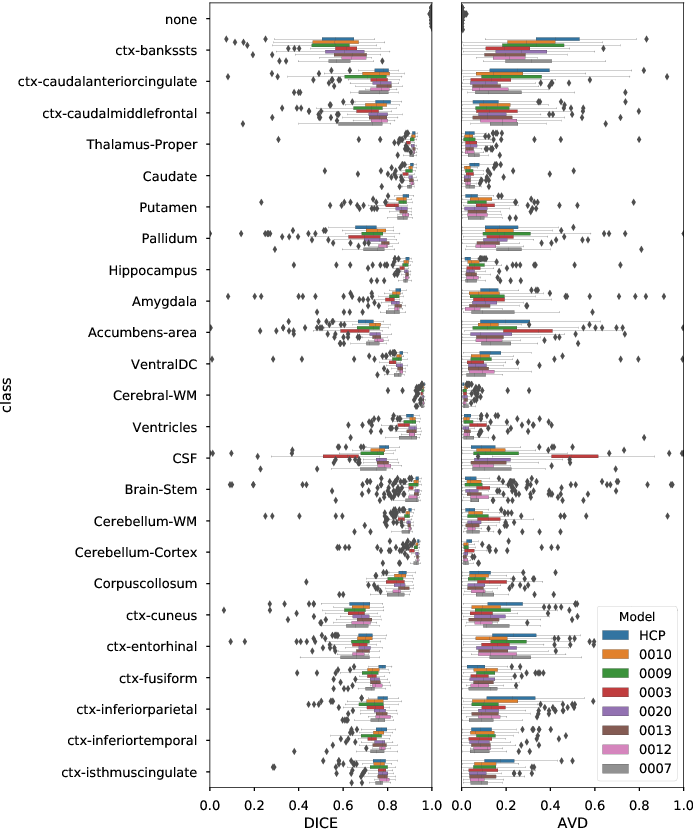}
\caption{DICE and AVD scores evaluated on $100$ testing subject from HCP and on $334$ testing subject by sites from FBIRN dataset for 25 regions from 50. DICE and AVD were measured on segmentation of whole MRI image $256\times  256\times  256$ with sampling of $1024$ $38\times  38\times 38$ subvolumes. For DICE to the right (higher values) is better and for AVD to the left (lower values) is better.} 
\label{fig:sites_part1}
\end{figure}

\begin{figure}[p]
\centering
\includegraphics[width=\textwidth, keepaspectratio]{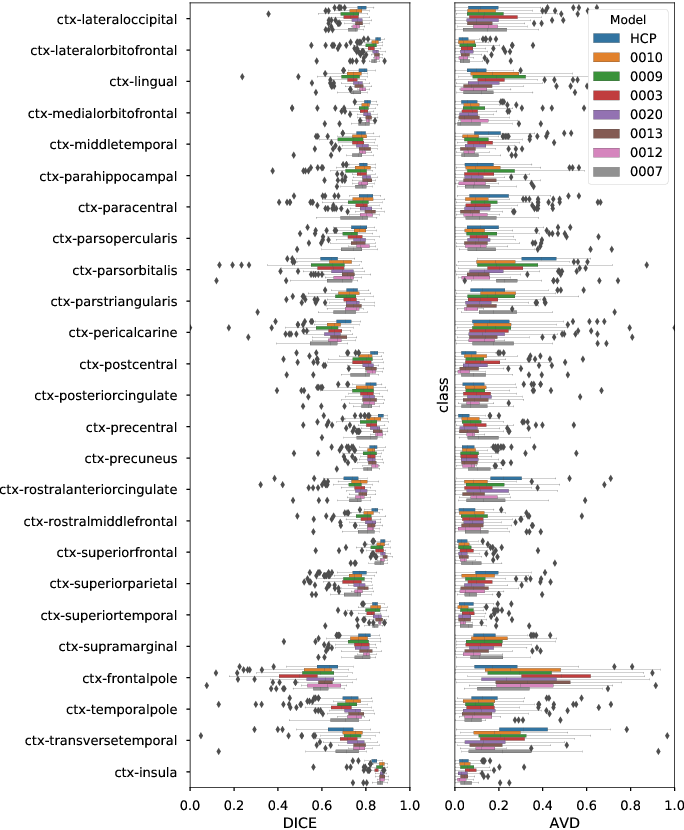}
\caption{DICE and AVD scores evaluated on $100$ testing subject from HCP and on $334$ testing subject by sites from FBIRN dataset for 25 regions from 50. DICE and AVD were measured on segmentation of whole MRI image $256\times  256\times  256$ with sampling of $1024$ $38\times  38\times 38$ subvolumes. For DICE to the right (higher values) is better and for AVD to the left (lower values) is better.}
\label{fig:sites_part2}
\end{figure}

Performance on $334$ testing subject from FBIRN dataset is shown in Figure~\ref{fig:fbirn_metrics_part1} and Figure~\ref{fig:fbirn_metrics_part2}. For comparing its performance we also show results on HCP dataset. Segmentation examples is shown in Figures~\ref{fig:fbirn_brains}. To remind FBIRN dataset has not just healthy subjects, but also patients with schizophrenia. Model shows similar performance by FBIRN dataset's sites in Figure~\ref{fig:sites_part1} and Figure~\ref{fig:sites_part2}.

\subsection{Statistical comparison}

\begin{figure}[p]
\centering
\includegraphics[width=\textwidth, keepaspectratio]{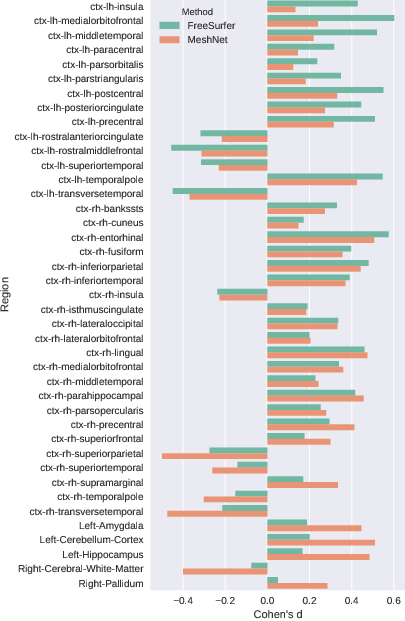}
\caption{$41$ ROIs with significant (w.r.t $< 0.05$) covariate $label$ and nonsignificant $method:label$.}
\label{fig:label_true_ml_false}
\end{figure}

\begin{figure}[h]
\centering
\includegraphics[width=\textwidth, keepaspectratio]{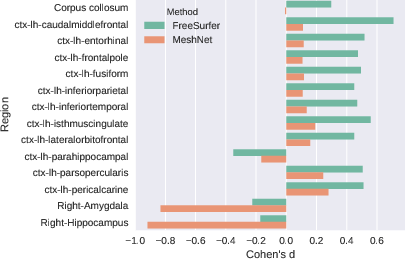}
\caption{$14$ ROIs with significant (w.r.t $< 0.05$) covariates $label$ and $method:label$.}
\label{fig:label_true_ml_true}
\end{figure}

\begin{figure}[p]
\centering
\includegraphics[width=\textwidth, keepaspectratio]{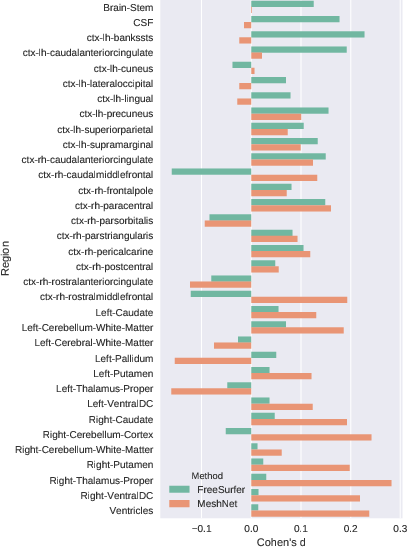}
\caption{$34$ ROIs with nonsignificant (w.r.t $< 0.05$) covariates $label$ and $method:label$.}
\label{fig:label_false_ml_false}
\end{figure}

\begin{figure}[h]
\centering
\includegraphics[width=\textwidth, keepaspectratio]{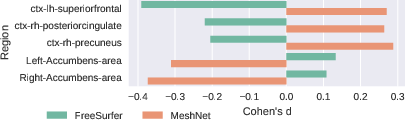}
\caption{$5$ ROIs with nonsignificant (w.r.t $< 0.05$) covariate $label$ and significant $method:label$}
\label{fig:label_false_ml_true}
\end{figure}

For statistical comparison, first, to prepare data we excluded from FBIRN dataset 14 subjects we had used for fine-tuning as training and validation set and subjects without gender. Then split most of ROIs on right and left hemispheres due disease could affect just one side of brain, calculated for each ROI's volume and whole brain volume as number of voxels. We ended up with $311$ subjects which are by gender $85$ females and $225$ males and by group $153$ healthy controls and $158$ patients with schizophrenia. Age is still between 18 to 62. Finally, we performed meta-analysis, described in Section~\ref{ssec:stat-meta}.

Repeated measurements ANOVA showed that in $41$ ROIs (Figure~\ref{fig:label_true_ml_false}) just $label$ is significant, therefore it doesn't matter which method is used and both methods explains group differences. In $14$ regions (Figure~\ref{fig:label_true_ml_true}) also interaction $method:label$ is significant, means methods have different slopes, but doesn't mean that both explain group differences. The covariates $label$ and $method:label$ aren't significant and therefore no differences are in $34$ regions (Figure~\ref{fig:label_false_ml_false}). In last $5$ regions (Figure~\ref{fig:label_false_ml_true}) the covariate $label$ isnt significant, but interaction $method:label$ is significant, means that slopes of method are intercepted. After analyzing Cohen'd we found that in $24$ regions Cohen'd of Meshnet is 2 times greater than for FreeSurfer and in $22$ regions otherwise. But in most cases of 94 regions Cohen's d has same direction. 

\section{Discussion}

\subsection{Overview}
In this paper, we investigated MeshNet as a tool for atlas segmentation of a human MRI brain image trained on a big dataset from the labels obtained using FreeSurfer. Proposed model shows outstanding speed for direct producing of atlas segmentation from raw data, comparable results in segmentation and most important that MeshNet is suitable solution for statistical meta-analysis. Furthermore, proposed model can utilize straightforward second modality T2 as input which leads to better segmentation quality.

Applied to FBIRN dataset MeshNet yeilded similar segmentation quality as on HCP dataset even we just used 7 brains from each site to fine-tune. It should be noted that FBIRN is harder dataset due different MRI machine settings in each site, containing MRI images with rotation, contrast and patients with schizophrenia. Therefore MeshNet is suitable for transfer learning solutions.

\subsection{Hyperparameters, training, testing and fine-tuning of Deep Neural Network}

To find best model of our network for atlas segmentation we have tried different number of features maps in each layer. From $51$, $65$, $71$, $81$ and $91$ feature maps per each layer $71$ was the optimal number getting the lowest cross-entropy loss. The weight initialization with xavier showed faster convergence and lower cross-entropy loss than with identity from~\cite{yu2015multi}. Batch normalization has showed lower cross-entropy loss if it is before activation function than if it's after, but it was conversely in gray matter and white matter segmentation task.

Training and testing was performed using nonoverlap and Gaussian distribution subvolume sampling. Nonoverlap sampling was used to be sure we cover whole brain. Gaussian distribution was used due brain is located mostly in the center of MRI image. The center of Gaussian distribution, first, was the middle of the volume $\{128, 128, 128\}$, but for better sampling, except model testing step, we used center mass based on the training set. It was located in $\{127, 145, 127\}$ which lowered our cross-entropy loss. For std of Gaussian distribution we chose $\{60, 60, 60\}$ due lower cross-entropy loss, before it was $\{50, 50, 50\}$. One training epoch was $30720$ subvolumes and model was validated on $27648$ subvolumes. For providing results in Section~\ref{ssec:qual_hcp} T1 model is $35$ epoch old and with dropout $p=0.125$ and $p=0.25$ are $77$ and $71$ epochs respectively, T1 and T2 model --- $35$, with  dropout $p=0.125$ --- $61$ and dropout $p=0.25$ --- $56$. Models with dropout takes more time to train, as expected.

To fine-tune for FBIRN dataset the weights of pretrained on HCP dataset network we run same pipeline for training on HCP, but we reduced learning step of Adam to $0.00001$. We didn't freezed any earlier layers, due dataset contains some rotations and brains of patients with schizophrenia. One fine-tuning epoch consisted from $7168 (7 \times 1024)$ subvolumes and same amount we used for validation. The FBIRN results were provided in Section~\ref{ssec:qual_fbirn} with model fine-tuned on $2243584$ subvolumes.

\section{Conclusions and future work}

Convolutional Neural Network is a very powerful tool that most likely can solve segmentation problem in neuroimaging but also reduce time needed for processing of large dataset. Therefore the purpose of our work was to create a uncomplicated as U-Net, powerful, generalizable from imperfect labeling, utilizing different neuroimaging modalities, fast and parameter-efficient model. The evidence we got through first work for gray matter and white matter segmentation and this continuation leads that MeshNet is capable for most our needs. Comparing MeshNet performance in statistical meta-analysis showed it as useful tool for large-scale studies. Next each step of modification need careful fine-tunning of hyperparameters and network architecture but the most important evaluating it by performing large-scale studies.

\section*{Acknowledgements}

This work was supported by NSF IIS-1318759 and NIH R01EB006841 grants.
Data were provided [in part] by the Human Connectome Project, WU-Minn Consortium (Principal Investigators: David Van Essen and Kamil Ugurbil; 1U54MH091657) funded by the 16 NIH Institutes and Centers that support the NIH Blueprint for Neuroscience Research; and by the McDonnell Center for Systems Neuroscience at Washington University. FBIRN dataset used for this study were downloaded from the Function BIRN Data Repository (http://fbirnbdr.birncommunity.org:8080/BDR/), supported by grants to the Function BIRN (U24-RR021992) Testbed funded by the National Center for Research Resources at the National Institutes of Health, U.S.A.

\section*{References}
\bibliography{references}
\end{document}